\documentclass[11pt]{article}
\usepackage[final]{acl}

\usepackage{times}
\usepackage{latexsym}
\usepackage{adjustbox}
\usepackage[T1]{fontenc}

\usepackage[utf8]{inputenc}
\usepackage{microtype}
\usepackage{inconsolata}
\usepackage{graphicx}

\usepackage{multirow} 
\usepackage{listings}
\usepackage{xcolor}
\usepackage{enumitem}
\usepackage{booktabs}
\usepackage[most]{tcolorbox}
\usepackage{listings}
\usepackage{bigfoot}

\newcommand\blfootnote[1]{%
  \begingroup
  \renewcommand\thefootnote{}\footnote{#1}%
  \addtocounter{footnote}{-1}%
  \endgroup
}

\title{Benchmarking Agentic Newswriting via Journalistic Workflows}

\author{Yen-Che Chien$^1\dagger$, Kuang-Da Wang$^1\dagger$, Wei-Yao Wang$^2\ddagger$, Wen-Chih Peng$^1$ \\
$^1$National Yang Ming Chiao Tung University, $^2$Sony Group Corporation \\
\texttt{dfg15243.cs12@nycu.edu.tw, gdwang.cs10@nycu.edu.tw} \\
\texttt{sf1638.cs05@nctu.edu.tw, wcpeng@cs.nycu.edu.tw}
}

\begin{document}
\maketitle
\begin{abstract}
Recent advances in autonomous digital agents from industry (e.g., Manus AI and Gemini’s research mode) highlight their potential for structured tasks through autonomous decision-making and task decomposition, but it remains unclear how well such systems support real-world information-intensive workflows.
We study this question in journalism, where newswriting requires iterative planning, contextual reasoning, and active discovery of missing background to produce a coherent article. We introduce NEWSAGENT, a benchmark for evaluating how agents search raw materials, select relevant information, and iteratively revise drafts through core journalistic functions.
Given a writing instruction and partial firsthand materials, agents must identify narrative perspectives, issue keyword-based queries, retrieve historical context, and generate complete news articles. Unlike typical summarization or retrieval tasks, essential context is not directly available and must be actively discovered, reflecting real-world reporting constraints. NEWSAGENT consists of 6k human-verified examples derived from real news. We evaluate open- and closed-sourced LLMs with commonly-used agentic frameworks on NEWSAGENT, which shows that agents are capable of retrieving relevant facts but struggling with planning and narrative integration. We believe that NEWSAGENT serves a realistic testbed for iterating and evaluating agent capabilities in terms of web data manipulation to real-world productivity. 
The benchmark resources are publicly available at \url{https://github.com/wywyWang/CoachAI-Projects}.

\end{abstract}
\blfootnote{$\dagger$ Both authors contributed equally to this research.}
\blfootnote{$\ddagger$ This work is independent of Sony Group Corporation.}

\section{Introduction}
Recently, modern computer-based applications increasingly rely on intelligent agents to conduct complex information-intensive reasoning tasks, benefiting users by automating real-world workflows and potentially improving productivity \citep{claude-computer-use}. 
From automated research assistants to search-based summarizers, systems such as Gemini’s deep research mode~\citep{geminideepresearch} and Manus AI~\citep{shen2025mindmachinerisemanus} demonstrate growing capabilities in retrieving, organizing, and synthesizing information from the web. 
These systems represent a shift from passive chatbots to interactive agents capable of planning and decision-making in open-ended environments~\citep{deps,react}.
\textit{This raises a central question: to what extent can modern agents execute realistic information-gathering and writing workflows beyond one-shot retrieval or summarization?}

\begin{figure}[t]
    \centering
    \includegraphics[width=\linewidth]{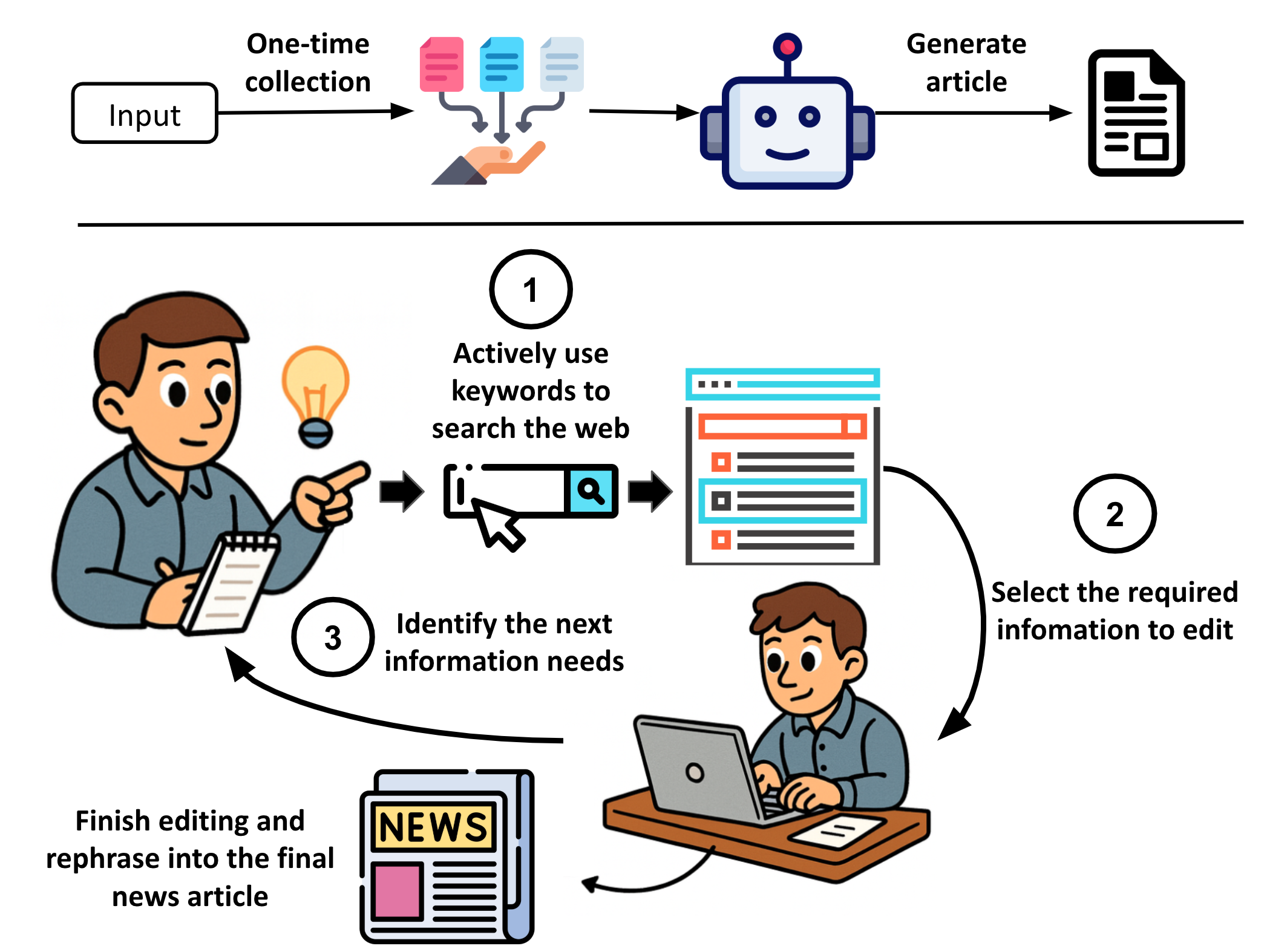}
    \caption{
    Comparison between one-time content generation (top) and the human journalistic workflow (bottom). While many automated tasks follow a one-time collection and generation process, human journalists start with limited firsthand data and iteratively search and add information to build a complete narrative. 
    }
    \label{fig:workflow}
\end{figure}

\begin{figure*}
    \centering
    \includegraphics[width=\linewidth]{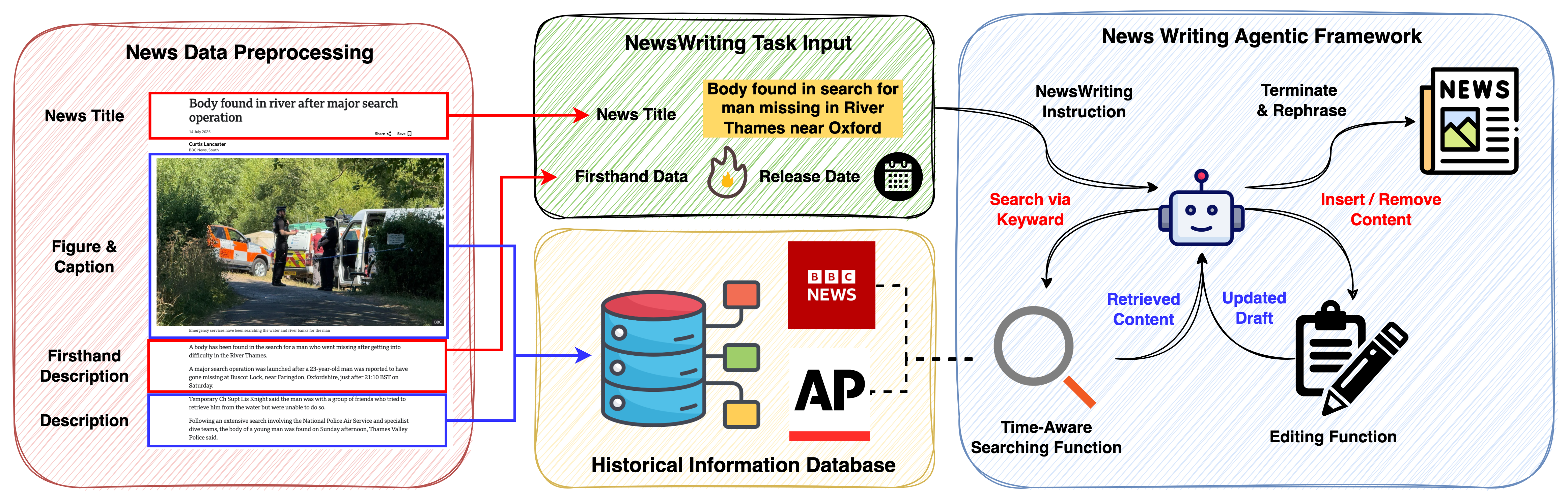}
    \caption{
    Overview of the \textbf{NEWSAGENT}. To construct the benchmark, we collect news articles, extract the news title and release date, where news articles are separated into firsthand information and historical information based on the published date. When performing news writing, the agent starts with the news title, release date, and firsthand information as the workflow input. In each task, the agent receives these inputs, searches the database for relevant context available before the release date, and decides how to edit the draft. Once the draft is complete, the agent rephrases it into the final news article, reflecting how human journalists gather information and refine stories through iterative editing.
    }
    \label{fig:overview}
\end{figure*}

Journalism offers an ideal testbed for exploring this question because newswriting is \emph{time-sensitive} and starts from \emph{incomplete firsthand material}, requiring agents to actively discover missing background and integrate evidence into a coherent, objective narrative. 
While existing agents excel in structured tasks such as program synthesis, summarization, and multi-step reasoning~\citep{cot,tot,ircot,kirag,wang2025talkstructurallyacthierarchically}, their capacity to emulate real-world reporting workflows remains relatively unexplored. 
In practice, journalists must identify newsworthy aspects, seek out historical evidence, and iteratively refine drafts as information needs emerge~\citep{Bloomberg,Bloomberg_seq,DBLP:conf/eacl/WangCP24}.
These requirements go beyond conventional retrieval-augmented generation~\citep{rag}, which typically relies on statically retrievable content and performs generation in a single pass.
As shown in Figure~\ref{fig:workflow}, journalistic writing is inherently iterative and exploratory, with planning and editorial refinement shaping what to search and what to include.

In this paper, we introduce NEWSAGENT\footnote{We use \textit{NEWSAGENT benchmark} for the dataset and task formulation, and \textit{NEWSAGENT pipeline} for the agentic search--edit--rephrase workflow used for evaluation.}, a benchmark for evaluating whether agents can execute journalistic workflows through autonomous searching and draft editing.

Prior work~\citep{llm_writers,AngleKindling} emphasizes narrative angle selection, but does not explicitly evaluate the workflow where agents must discover missing context and iteratively revise drafts under temporal constraints.
To model this setting, NEWSAGENT provides two core functions: a \textbf{time-aware search function} for retrieving historical context before the release date, and an \textbf{editing function} that incrementally inserts and removes content.
Given a news title, the release date, and partial firsthand materials (e.g., transcripts or descriptions), an agent issues queries, gathers relevant background, and refines a draft before rephrasing it into a final news article (Figure~\ref{fig:overview}).

Our benchmark contains 6,237 human-verified examples from real-world news events.
We evaluate both closed-source and open-source LLMs, normalizing all inputs into textual representations to focus on agentic newswriting at the semantic level.
We benchmark agents at two complementary levels.
\textbf{Function-wise} evaluation computes F1 against reference (human-written) articles to measure whether an agent's retrieval and edit decisions \emph{align with} choices made by past human journalists (i.e., similarity in selected background facts), rather than treating the reference as the only correct story; accordingly, higher function-wise scores are \emph{not} guaranteed to imply better newswriting quality.
\textbf{End-to-end} evaluation measures overall newswriting performance via a dimension-wise GPT-4~\citep{gpt4} comparative protocol across six journalistic dimensions: Factuality, Logical Consistency, Importance, Readability, Objectivity, and Journalistic Style, where higher scores indicate better quality; we additionally validate the protocol with human judgments to quantify agreement.
Together, this separation enables both diagnostic analysis of evidence-selection behavior and reliable assessment of writing quality.

Our efforts are summarized as follows:
\begin{itemize}
    \item 
    We construct 6,237 human-verified examples and formalize newswriting as an agentic process that starts from incomplete firsthand materials and requires actively discovering missing background under a release-date constraint.
    \item 
    NEWSAGENT evaluates explicit \texttt{Search} and \texttt{Edit} decisions, enabling analysis of how agents retrieve, select, and revise evidence during drafting.
    \item 
    We show that alignment with human evidence selection does not necessarily translate to better end-to-end newswriting, motivating future work on planning and narrative integration beyond matching references.
\end{itemize}

\section{Related Work}
Language models have been explored for supporting narrative generation in various domains of journalism. 
\citet{plan_and_write} proposed the Plan-and-Write framework, which first constructs a storyline plan and then generates a narrative, demonstrating that explicit planning improves coherence and relevance in automatic storytelling. 
\citet{AngleKindling} introduced AngleKindling, a system that supports journalistic angle ideation from press releases by suggesting editorial framings and summarizing key points, highlighting LLMs' potential in early-stage editorial decision-making. 
\citet{llm_writers} examined whether LLMs plan like human writers by comparing LLM-generated coverage of press releases with that of professional journalists, revealing differences in angle selection, fact use, and narrative focus. 
\citet{Bloomberg_seq} presented Sequentially Controlled Text Generation, a production system for Bloomberg journalists that decomposes article writing into ordered generation stages, enabling more controllable and accurate financial news.
In retrieval-related tasks, there has been a shift from static one-time retrieval toward more interactive, multi-step retrieval processes. 
\citet{ircot} proposed IRCoT, interleaving retrieval with step-by-step reasoning for multi-step QA, improving both retrieval and answer accuracy.
\citet{kirag} introduced KiRAG, a knowledge-driven iterative retriever that decomposes documents into knowledge triples and dynamically retrieves relevant triples to adapt to evolving information needs, achieving substantial gains in multi-hop QA.
These works reflect a growing trend toward making content generation more interactive, moving beyond one-time retrieval and generation to iterative retrieval, reasoning, planning, and generation. 
However, they often diverge from how journalists produce news, where information is acquired under constraints: key facts may be missing, require targeted searches or verification, and new angles can demand fresh sources mid-process. Many systems assume all material is available upfront, overlooking the investigative and evolving nature of reporting. 

\subsection{LLMs as Agents}
Our work builds on recent advances in enabling large language models to act as agents that reason, plan, and take actions in interactive environments. Chain-of-Thought (CoT)~\citep{cot} promotes step-by-step reasoning, ReAct\citep{react} integrates reasoning with external actions, Reflexion~\citep{reflexion} enables self-improvement through reflection, and Tree-of-Thought (ToT)~\citep{tot} supports branching and backtracking in reasoning. Benchmarks such as AgentBench~\citep{agentbench} evaluate agent performance across web navigation, tool use, and knowledge-intensive tasks, while MMSearch~\citep{mmsearch} focuses on multimodal retrieval and summarization. 
Although MMSearch is topically related, it focuses on one-shot retrieval and generation, where all context is gathered in a single step. In contrast, NEWSAGENT models newswriting as an iterative process with evolving information needs, reflecting the dynamic editorial workflow of real-world journalism.
\section{NEWSAGENT}
In this section, we present the NEWSAGENT benchmark and the NEWSAGENT pipeline for evaluating whether agents can perform core journalistic workflows. 
We first describe the dataset curation process (Section~\ref{sec:newsagent_instruction}), then detail the agentic pipeline (Section~\ref{sec:newsagent_pipeline}), and finally outline the evaluation protocol (Section~\ref{sec:newsagent_eval}).

\subsection{Dataset Curation for Newswriting}
\label{sec:newsagent_instruction}
To construct realistic and diverse tasks, we curated a large corpus of real-world news articles from BBC\footnote{\url{https://www.bbc.com/}} and APNews\footnote{\url{https://apnews.com/}}, covering diverse domains such as politics, sports, technology, and science. The initial collection comprised 31{,}097 articles published between 1 June 2025 and 14 July 2025. We removed non-English content and non-text formats such as videos, audio streams, and live updates.

\subsubsection{Object definition}
The unit of content in NEWSAGENT is an \textit{object}, defined as a semantically coherent piece of information represented in text form. All objects follow a unified JSON structure:
\begin{itemize}
    \item \textbf{Description:} Sentences directly from the article body, e.g., \texttt{\{"text": "Gray's report criticized leadership at No.10."\}}.
    \item \textbf{Caption:} Image captions provided in the source webpage, prefixed with \texttt{[Caption]} to distinguish them from other textual sources.
    \item \textbf{Quote/Transcript:} Speaker-attributed speech extracted from the article content, stored in the format \texttt{[Speaker's Name] content}.
\end{itemize}
Both the \textit{firsthand information database} and the \textit{historical information database} adopt this unified text-object format, ensuring consistent semantic representation across heterogeneous newsroom sources.

\subsubsection{Extraction and verification}
\label{subsec:reference-selected evidence}
We used GPT-4 to classify each object as \textit{firsthand} or \textit{historical}. Firsthand data includes event descriptions, direct quotations, captions, and transcripts available to a journalist at the time of publication. Historical information includes earlier developments, background context, and retrospective references. To prevent hallucinations, we programmatically verified that each extracted object appeared verbatim in the original article. Articles with more than five extraction failures or without any historical information were discarded. Two annotators independently reviewed all GPT-4 classifications, and any object without agreement was removed, resulting in 6{,}237 validated articles.

For evaluation purposes, the objects that remain traceable to the human-written article constitute the \textit{reference-selected evidence}. That is, the reference evidence is not separately annotated by an additional journalist; rather, it is derived from the set of verified objects that are actually included in the human-written reference article. We make this distinction explicit because the function-wise evaluation compares agent-selected evidence against this reference-evidence set.

\subsubsection{Dataset Distribution}
Across the dataset, 69\% of objects are classified as firsthand information and 31\% as historical information.
Descriptions constitute the largest share, while captions and transcripts form smaller subsets. Notably, most captions and transcripts are labeled as firsthand information, reflecting typical newsroom practice: such materials often contain valuable, sometimes exclusive, content that serves as a core foundation for reporting.

\subsubsection{Task formulation}
Each validated article is converted into a newswriting task in which the agent is provided with the news title, release date, and all firsthand information. The agent must produce a complete article through iterative search, editing, and rephrasing, where the \textit{object} serves as the smallest unit for these operations. Historical information is stored separately with timestamps, enabling retrieval during search while preventing access to content published after the release date. All inputs are represented as text objects (including captions and transcripts), so NEWSAGENT evaluates agentic newswriting at the semantic level.

\subsection{NEWSAGENT Pipeline}
\label{sec:newsagent_pipeline}

Figure~\ref{fig:overview} illustrates the NEWSAGENT workflow as an iterative observation--action (reasoning--action) loop. At each step, the agent receives an observation, selects an action, and updates the current draft accordingly. The process continues until the agent issues a \textbf{Terminate} action, after which the final draft is rephrased into a publishable news article.

\subsubsection{Observation}
At each iteration, the agent observes:
\textbf{(1) Draft state}: All content currently included in the article draft, representing accumulated information deemed relevant to the title.
\textbf{(2) Task inputs}: The news title, simulated release date, and firsthand information of the target article.
\textbf{(3) Retrieved content}: Objects returned from previous \texttt{Search} actions, drawn from the historical information database.
\textbf{(4) Operation message}: Feedback from previous action. This returns a success message if the action is valid, or an error message in one of the following cases:
(i) search yields no results;
(ii) insert attempts to use content not in the retrieved set;
(iii) remove targets content not present in the draft state;
(iv) the action cannot be parsed due to invalid formatting or hallucinations.

\subsubsection{Actions}
We frame newswriting as a structured agentic process with two fundamental capabilities:
(1) a \textbf{time-aware search function} for retrieving historical content, and
(2) an \textbf{editing function} for modifying the draft.
These are instantiated via three concrete actions:
\begin{enumerate}
\item \textbf{Search}: Generate a keyword query and retrieve historical content published strictly before the simulated release date, preventing access to future information. The search function returns the top-$k$ results ($k=5$ in our experiments) with cosine similarity above $0.7$. We choose these values as a conservative trade-off between coverage and noise under a fixed tool budget.
\item \textbf{Insert}: Add selected retrieved objects to the draft, enriching factual and contextual coverage. Only objects returned by prior \texttt{Search} actions are allowed; attempts to insert other content trigger an error message.
\item \textbf{Remove}: Delete existing objects from the draft state. Only objects already present in the draft can be removed.
\end{enumerate}
The agent alternates between these actions within the observation--action loop, progressively refining the draft until it deems the article complete.

\subsubsection{Termination and Rephrasing}
When the agent issues a \textbf{Terminate} action, the current draft---represented as a sequence of selected objects---is passed to a \textbf{rephrasing} step.
Importantly, the draft is not yet a publishable article: a good news story requires narrative structure (e.g., a lead, transitions, emphasis, and wrap-up) and editorial framing (e.g., what to foreground and how to connect evidence), rather than merely listing facts.
The rephrasing step \emph{realizes} these discourse-level elements by converting the selected objects into fluent paragraphs with appropriate cohesion and attribution.
To reduce unsupported content during rephrasing, we apply a post-hoc traceability check.
Rephrased text is compared against the selected objects using the same semantic retrieval mechanism as the search function; if no corresponding evidence can be retrieved, the rephrasing step is re-executed.

\begin{tcolorbox}[colback=gray!10,colframe=gray!40,boxrule=0.5pt,arc=2pt,left=6pt,right=6pt,top=6pt,bottom=6pt]
\textbf{Why this design matters.}
NEWSAGENT separates two sources of variation in newswriting: (i) \emph{evidence selection} under temporal constraints (what the agent searches for and keeps in the draft), and (ii) \emph{narrative realization} (how the selected evidence is organized and written into a coherent story).
Since journalistic narratives are inherently non-unique, similarity to a reference article is informative as \emph{alignment} to past editorial choices but is not itself the final notion of quality.
This motivates our two-level evaluation below: function-wise metrics diagnose alignment in evidence acquisition/retention, while end-to-end evaluation measures overall newswriting performance.
\end{tcolorbox}

\subsection{Evaluation Protocol}
\label{sec:newsagent_eval}
NEWSAGENT evaluates agent performance from two complementary perspectives: diagnostic process alignment and end-to-end newswriting quality.

\paragraph{Function-wise metrics (diagnostic alignment).}
We assess each core function: time-aware searching and draft editing. Here, the \textit{reference-selected evidence} refers to the verified text objects derived from the human-written reference article, as described in Section~\ref{subsec:reference-selected evidence}. For \texttt{Search}, we compare retrieved objects against this reference-evidence set and compute Precision, Recall, and F1 to quantify how similarly the agent retrieves background information relative to past human editorial choices under the same temporal constraint. For \texttt{Edit}, we condition on reference-selected evidence being present in the agent’s observation and measure whether it is retained in the draft after editing actions (\texttt{Insert} and \texttt{Remove}), again reporting Precision, Recall, and F1.
We emphasize that these scores are diagnostic alignment measures rather than end-to-end quality measures: a higher function-wise F1 is not inherently better, but instead indicates closer alignment to the subset of evidence selected in the human-written reference article. Since journalistic writing is open-ended and non-unique, divergence from the reference evidence does not necessarily imply lower article quality; it may also reflect a different but still valid narrative angle or evidence-selection strategy.

\paragraph{End-to-end metrics (newswriting performance).}
In contrast, end-to-end evaluation directly measures overall article quality after the full search--edit--rephrase process, where higher preference indicates better newswriting performance. We design a dimension-wise comparative evaluation framework in which an LLM judge compares two candidate articles across six dimensions: factual consistency, logical consistency, importance, readability, objectivity, and journalistic style. Based on the human validation results described below, we use GPT-4 as the primary judge in the main end-to-end evaluation, while also verifying the robustness of the protocol with additional judge models from a different model family.
For each dimension, it outputs a preference with brief justification, and then synthesizes these judgments into an overall preference decision. The order of candidate articles is randomized to mitigate positional bias.
The exact dimension-wise evaluation prompt is provided in Appendix~\ref{sec:judge_prompt}.

\paragraph{Validation.}
To assess the reliability of the LLM-based evaluation protocol, we randomly sampled 100 human-written news articles and used GPT-4o and GPT-4o-mini to generate corresponding articles, yielding three candidate articles per news event: one human-written article and two model-generated articles. We then constructed all pairwise comparisons among the three candidates, resulting in 300 article pairs in total, and collected human preference labels under an anonymized and randomized setup.

After excluding 17 pairwise comparisons labeled as ties by human annotators, 283 comparisons were retained for agreement analysis. We evaluated these same pairs using both standard single-turn judging and our dimension-wise comparative protocol. For GPT-4, standard single-turn evaluation achieved 53\% agreement with human judgments, whereas our dimension-wise evaluation achieved 72\%. We observed the same pattern with Gemini-2.5-Flash: standard single-turn evaluation achieved 59\% agreement, while the dimension-wise protocol improved this to 69\%. These results indicate that the dimension-wise protocol better aligns with human preferences than single-turn holistic judging across judge families.

Since GPT-4 showed the strongest overall performance under our validation setup, we use GPT-4 as the primary judge for the main end-to-end results reported in this paper. Details of the sampling procedure, annotation setup, and agreement computation are provided in Appendix~\ref{sec:human_validation}.
\section{Experiments}
\subsection{Experimental Setup}
\subsubsection{Frameworks}
We adopt the ReAct framework~\citep{react} as the agentic backbone for our evaluation because it explicitly exposes the interaction between reasoning and action selection.
By interleaving intermediate reasoning with concrete operations such as \texttt{Search}, \texttt{Insert}, and \texttt{Remove}, ReAct allows us to observe not only final outputs but also \emph{when} an agent chooses to act and \emph{which} action it selects, which is central to evaluating iterative newswriting workflows.
ReAct has been widely adopted in agentic reasoning research~\citep{reflexion}, making it a well-understood and representative framework for assessing agent behavior in open-ended tasks.

Our setting requires fine-grained control over \texttt{Search} and \texttt{Edit} operations with strict execution formats.
Models vary substantially in their ability to reliably follow structured action specifications, which can obscure whether failures stem from decision-making limitations or from action-formatting errors~\citep{DBLP:journals/corr/abs-2505-20139}.
To disentangle these factors and ensure consistent evaluation conditions, we implement two execution modes within ReAct:
\begin{itemize}
\item \textbf{1-step setting}: The agent directly executes the selected operation (\texttt{Search}, \texttt{Insert}, or \texttt{Remove}) in a single step, providing the complete query or content.
\item \textbf{2-step setting}: The agent first selects the operation type and, in a subsequent step, specifies the query or content to be operated on.
\end{itemize}

The 2-step design reduces action-specification burden and isolates action selection from execution. 
All models share the same prompt and the action schema.
The prompt template is provided in Appendix~\ref{sec:agent_prompt}.

\begin{table}[t]
    \centering
    \caption{List of LLMs evaluated for newswriting capability, with release dates.}
    \begin{adjustbox}{width=\columnwidth}
    \begin{tabular}{lc}
    \toprule
    \textbf{Model ID} & \textbf{Release Date} \\
    \midrule
    \texttt{gpt-4o-2024-11-20}~\citep{gpt_4o} & 2024-11-20 \\
    \texttt{gpt-4o-mini-2024-07-18}~\citep{gpt_4o} & 2024-07-18 \\
    \texttt{gemma-3-27b-it}~\citep{gemma3} & 2025-03-12 \\
    \texttt{Qwen3-32B}~\citep{qwen3} & 2025-04-29 \\
    \texttt{Llama-4-Scout-17B-16E-Instruct}~\citep{llama4scout} & 2025-04-05 \\
    \bottomrule
    \end{tabular}
    \end{adjustbox}
    \label{tab:model_list}
\end{table}

\subsubsection{Models}
We evaluate both closed-source and open-source LLMs on NEWSAGENT. Table~\ref{tab:model_list} lists the models along with their model IDs and release dates.
To contextualize agent performance, we include a rule-based baseline as a mechanical lower bound. This baseline performs no planning or draft revision: it retrieves historical objects using the same embedding-based search procedure as our system and then inserts the top five retrieved results into the draft without further editing or rephrasing, directly outputting the resulting draft as the final article.
This baseline isolates the value of agentic planning and editorial integration beyond retrieval alone.

\paragraph{Data Leakage Avoidance}
Table~\ref{tab:model_list} reports the release dates of all evaluated models to ensure that none were trained on data beyond June 2025. Specifically, there is no overlap with our dataset. This temporal gap ensures that measured performance reflects agentic capabilities rather than memorization of target articles.

\subsubsection{Implementation Details}
When the agent issues a \texttt{Search} operation, the system computes cosine similarity scores between the query and all entries in the historical database using text embeddings from \texttt{all-MiniLM-L6-v2}\footnote{\url{https://huggingface.co/sentence-transformers/all-MiniLM-L6-v2}}. The top five results with similarity greater than 0.7 are returned to the agent.
For \texttt{Insert} and \texttt{Remove} operations, the system verifies that the target object exists in the most recent search results or the current draft. Matching requires more than semantic similarity: after removing punctuation, the text must match exactly.
This design keeps draft editing verifiable at the object level and avoids ambiguous credit assignment from paraphrased insertions.
If the object is not found, an error is returned and the operation is still counted toward the total operation limit.
The agent may issue \texttt{Terminate} at any step; we additionally impose a cap of 20 operations (\texttt{Search}, \texttt{Insert}, or \texttt{Remove}) to standardize tool budgets and prevent degenerate loops.
All model parameters are kept at default settings. Closed-source models are accessed via the OpenAI API\footnote{\url{https://platform.openai.com/docs/api-reference}}. Open-source models are served through the DeepInfra API\footnote{\url{https://deepinfra.com/docs}}.
We report operation budget and decoding configuration in Appendix~\ref{sec:exec_settings}.

\begin{table*}[t]
\centering
\caption{Precision, Recall, and F1 scores for searching and editing function operations under the 1-step and 2-step execution settings. Dashes (-) indicate that the model could not perform the function in the 1-step or 2-step setting. The highest score in each column is shown in bold. Scores are computed at the article level and macro-averaged across the evaluation set; the average F1 is not equal to the F1 computed from the average Precision and Recall.
}
\label{tab:prec_rec_f1_models}
\small
\renewcommand{\arraystretch}{1.4}
\setlength{\tabcolsep}{5pt}
\begin{adjustbox}{width=\textwidth}
\begin{tabular}{l|ccc|ccc|ccc|ccc}
\toprule
 & \multicolumn{6}{c|}{\textbf{Searching}} & \multicolumn{6}{c}{\textbf{Editing}} \\
\cline{2-13}
 & \multicolumn{3}{c|}{1-step} & \multicolumn{3}{c|}{2-step}
 & \multicolumn{3}{c|}{1-step} & \multicolumn{3}{c}{2-step} \\
\cline{2-13}
 & Prec. & Rec. & F1 & Prec. & Rec. & F1 & Prec. & Rec. & F1 & Prec. & Rec. & F1 \\
\midrule
\textbf{GPT-4o} & \textbf{0.327} & 0.292 & \textbf{0.233} & 0.635 & \textbf{0.095} & \textbf{0.150} & \textbf{0.808} & \textbf{0.208} & \textbf{0.267} & 0.847 & 0.083 & 0.147 \\
\textbf{GPT-4o mini} & 0.282 & \textbf{0.322} & 0.231 & 0.700 & 0.094 & 0.140 & 0.692 & 0.166 & 0.237 & 0.845 & 0.089 & 0.142 \\
\midrule
\textbf{Gemma-3-27b-it} & 0.219 & 0.321 & 0.206 & 0.732 & 0.086 & 0.142 & 0.653 & 0.179 & 0.214 & 0.861 & \textbf{0.092} & \textbf{0.152} \\
\textbf{Qwen3-32B} & -- & -- & -- & \textbf{0.844} & 0.071 & 0.120 & -- & -- & -- & 0.843 & 0.071 & 0.121 \\
\textbf{Llama-4-Scout-17B-16E-Instruct} & -- & -- & -- & 0.837 & 0.079 & 0.126 & -- & -- & -- & \textbf{0.941} & 0.075 & 0.126 \\
\midrule
\textbf{Rule-Based} & 0.040 & 0.174 & 0.058 & -- & -- & -- & 0.040 & 0.174 & 0.058 & -- & -- & -- \\
\bottomrule
\end{tabular}
\end{adjustbox}
\end{table*}

\subsection{Experimental Analysis} \label{sec:main_result}
\subsubsection{Agents Select Different Evidence Than Human Journalists}
Table~\ref{tab:prec_rec_f1_models} reports precision, recall, and F1 scores for \texttt{Search} and \texttt{Edit} under both 1-step and 2-step execution settings, measuring alignment between agent-selected evidence and the content chosen by human journalists.
Across all models, F1 scores are consistently low, indicating substantial divergence in evidence selection even when relevant information is available in the historical database.
Importantly, this divergence does not imply lower article quality; rather, it reflects that agents and human journalists often prioritize different subsets of information when constructing a news narrative. Comparing execution modes, the 2-step setting generally increases precision but lowers F1.
This suggests that decomposing action execution leads agents to select a smaller, more conservative set of evidence aligned with human choices, while omitting additional contextual information that agents might otherwise use to build richer narratives.
We further analyze operation counts in Appendix~\ref{sec:tool_use_analysis}.

\begin{figure*}[h]
    \centering
    \includegraphics[width=\linewidth]{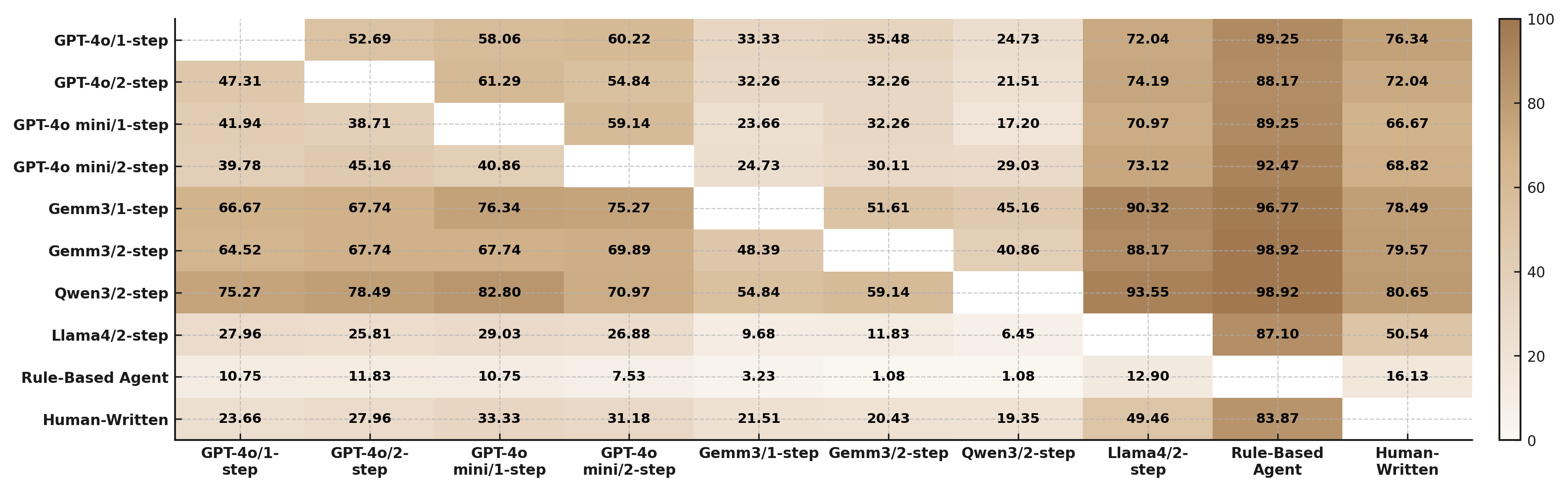}
    \caption{
    Pairwise head-to-head win rates (\%) for the end-to-end newswriting task.
    Each cell represents the percentage of cases where the row model outperformed the column model. Darker cells indicate higher win rates; rows with consistently darker shading identify models that robustly outperform multiple baselines.
    }
    \label{fig:pairwise_win_rates_generated}
\end{figure*}

\subsubsection{Closed-Source Is Not Always Best}
Figure~\ref{fig:pairwise_win_rates_generated} presents the pairwise head-to-head win rates for the end-to-end newswriting task. The results show that closed-source LLMs such as GPT-4o and GPT-4o mini do not consistently outperform high-performing open-source models such as Qwen3-32B and Gemma-3-27b-it. This challenges the common assumption that greater general-purpose reasoning capability necessarily translates into superior performance in targeted editorial workflows. Since the NEWSAGENT benchmark emphasizes focused search and iterative editing rather than complex long-horizon reasoning, higher reasoning capacity does not always yield higher-quality final articles.

Human-written articles also do not achieve the highest win rates. This is consistent with our earlier finding that greater F1 alignment with human-selected content does not guarantee better end-to-end performance, indicating that high-quality narratives can emerge from information selections that differ from human editorial choices.
The 2-step setting further illustrates this disconnect: although it often improves precision and alignment with human evidence selection, it does not reliably improve win rates and can even reduce performance for models such as GPT-4o and Gemma-3.

\begin{figure*}[t]
    \centering
    \includegraphics[width=\linewidth,trim={3cm 0cm 3cm 0cm},clip]{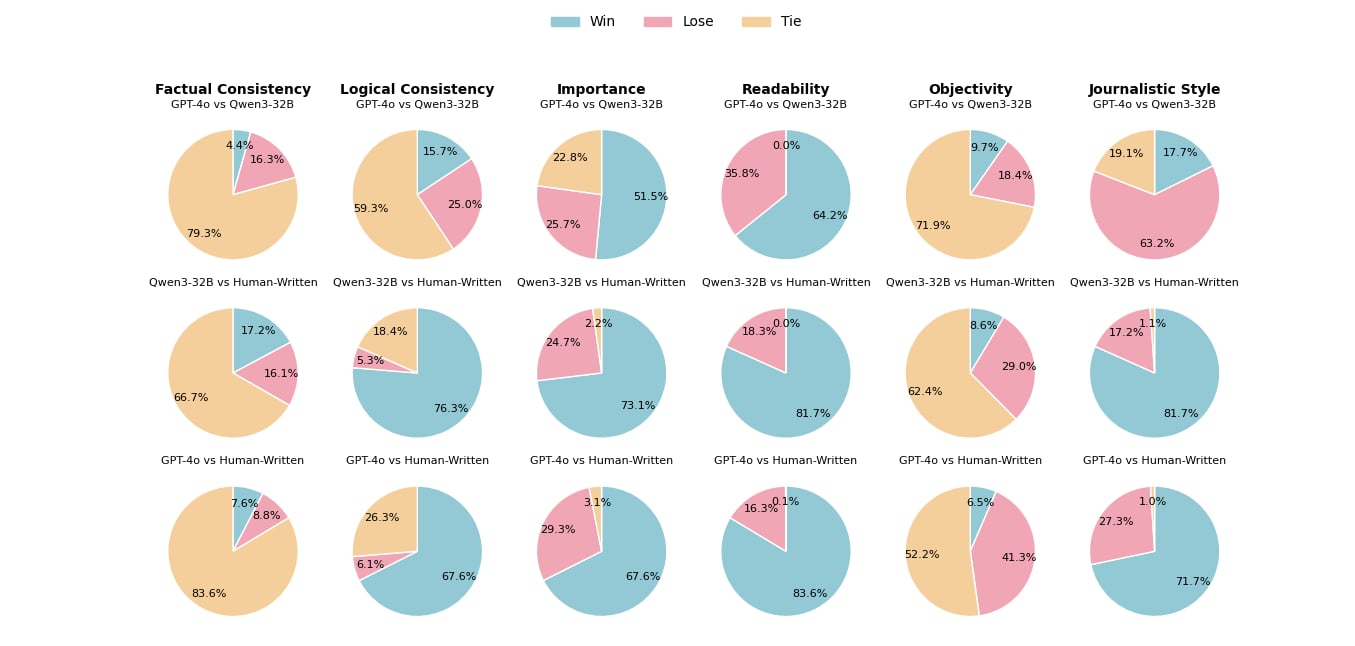}
    \caption{
    Dimension-wise preference distributions for pairwise model comparisons across six evaluation dimensions.
    Each pie chart shows the proportion of wins, losses, and ties for the model named in the chart title.
    }
    \label{fig:dimension_breakdown}
\end{figure*}

\subsubsection{Where Models Win and Lose} \label{sec:error}
To characterize model strengths, we conduct a dimension-wise analysis using our GPT-4 evaluation across six dimensions. We focus on two representative systems: GPT-4o (1-step), the strongest closed-source model in our end-to-end comparisons, and Qwen3-32B (2-step), the strongest open-source model. Both are also compared against human-written articles. Figure~\ref{fig:dimension_breakdown} highlights where wins and losses concentrate, separating global performance from dimension-specific trade-offs.

\paragraph{Qwen3-32B wins on journalistic style, while GPT-4o leads on readability.}
GPT-4o’s clearest advantage is \textit{Readability}, producing fluent and easy-to-follow narratives. In contrast, Qwen3-32B shows a strong edge in \textit{Journalistic Style} and also performs well on \textit{Importance}, which together drive its higher overall win rates in Figure~\ref{fig:pairwise_win_rates_generated}.

\paragraph{Human-written articles prioritize accuracy and brevity.}
Human-written articles are competitive on \textit{Factual Consistency} and \textit{Objectivity}, often resulting in ties, but they tend to be more concise and include less extended background. In contrast, Qwen3-32B incorporates more additional historical context beyond what appears in the reference article and uses it to strengthen narrative continuity, which correlates with higher ratings in \textit{Journalistic Style} and \textit{Importance}. 
A qualitative case study illustrating these contrasting evidence selection strategies is provided in Appendix~\ref{sec:case_study}.

\section{Conclusion}
This paper introduces \textsc{NEWSAGENT}, a benchmark for evaluating agentic newswriting under realistic journalistic constraints.
NEWSAGENT models newswriting as an iterative process in which agents must search for background information, select and revise evidence, and produce a complete article given incomplete firsthand materials and a fixed release date.
Unlike prior benchmarks that emphasize one-shot generation from pre-collected data, NEWSAGENT evaluates multi-step article construction that mirrors real newsroom workflows.
The benchmark contains over 6k news articles and centers on journalistic functions such as searching, inserting, removing, and rephrasing information.
Experiments across closed- and open-source models show that similarity to human-written articles is not a reliable indicator of quality, and that open-source models such as Qwen3-32B can match or exceed closed-source models on specific journalistic dimensions.
Overall, \textsc{NEWSAGENT} provides a practical testbed for studying agentic planning and editorial integration in newswriting.

\section*{Limitations}
NEWSAGENT is constructed from English-language articles drawn from two publishers within a limited time window. As a result, the benchmark does not fully cover the diversity of editorial styles, reporting practices, or topic distributions found in broader news ecosystems.
In addition, NEWSAGENT uses a fixed retrieval and controlled tool interface to support reproducible evaluation. This makes agent behavior easier to compare, but it does not fully capture real-world web reporting settings, where systems must also handle ranking noise, heterogeneous source quality, duplication, freshness, and source credibility.
Finally, our end-to-end evaluation relies on LLM-based judging. Although we strengthen this protocol with human validation and validation across judge families, any automatic judge may still exhibit residual preferences for particular writing styles or dimensions.

\section*{Ethical considerations}
NEWSAGENT is built from publicly available news articles and is intended to study agentic newswriting workflows rather than to replace professional journalists. As with any system that generates news-like text, there is a risk of misuse for producing misleading, plagiarized, or low-quality content if deployed without human oversight. We recommend that applications of agentic newswriting include clear disclosure of AI assistance, editorial review, and safeguards for sensitive topics.

\section*{Acknowledgments}
This work was supported by the National Science and Technology Council (NSTC) of Taiwan under Grants 113-2425-H-A49-001.

\bibliography{custom}
\appendix

\begin{figure*}[ht]
    \centering
    \includegraphics[width=\linewidth]{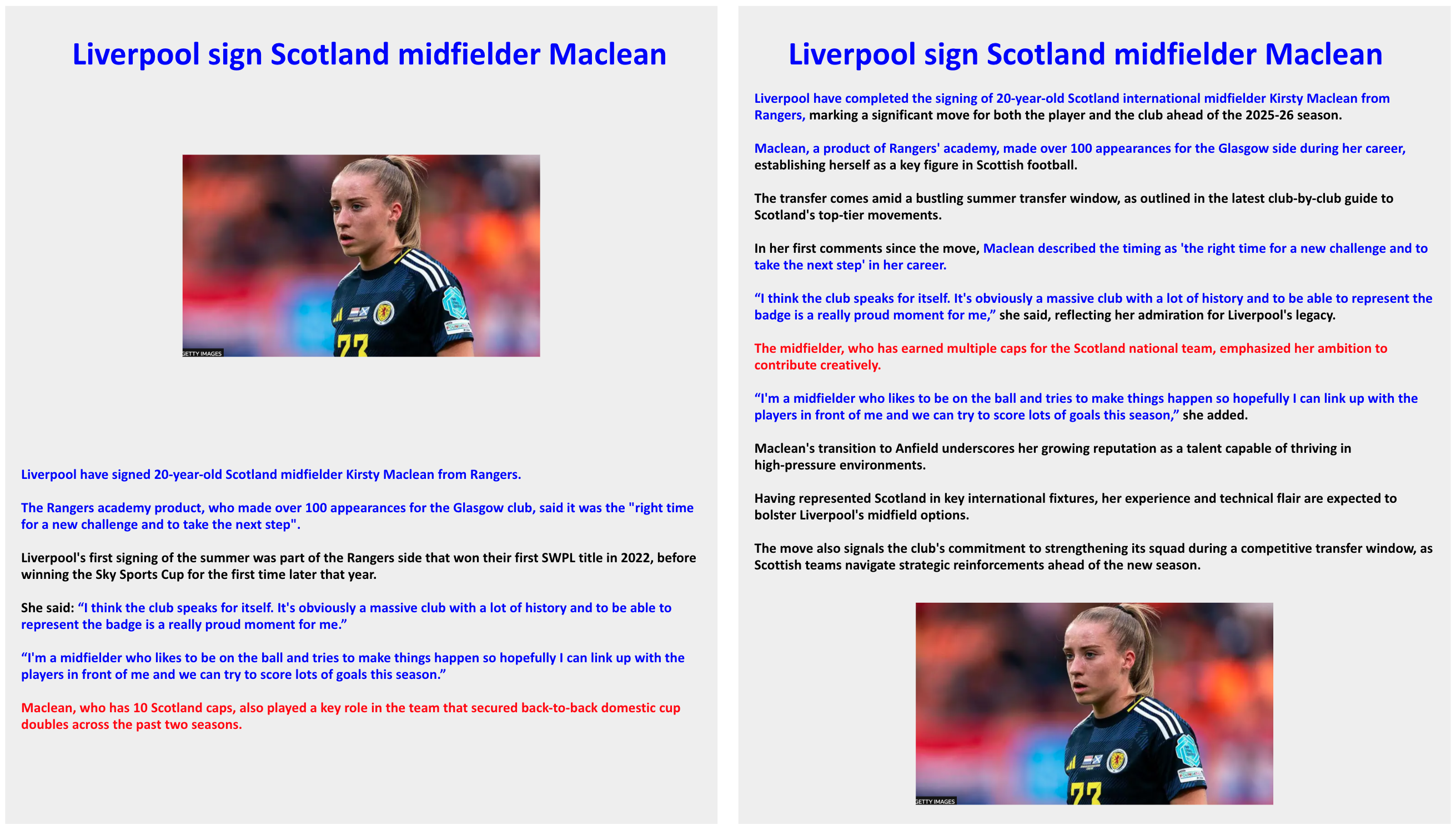}
    \caption{
    Comparison between a BBC human-written article (left) and Qwen3-32B output (right) for the same news event. Blue text denotes firsthand information, including image captions, unique to each version. Red text marks historical information shared by both but used in different ways. Black text represents content exclusive to one version. The human-written article focuses on concise factual delivery, whereas Qwen3-32B integrates a broader set of historical details to enhance narrative continuity and stylistic richness. The original BBC article is available at \url{https://www.bbc.com/sport/football/articles/c5y78338ylyo}.
    }
    \label{fig:qwen3_vs_human_example}
\end{figure*}

\section{Additional Analyses}
\label{sec:additional_analysis}


\begin{table*}[h]
\centering
\caption{
Function usage statistics and average token counts. 
\textit{Insert Fail} indicates the average number of insert attempts on content not present in the search results. 
Searching are allowed to return no results, so search failure counts are not reported. 
Dashes (-) indicate that the model could not perform the function in 1-step setting.
}
\label{tab:count_models}
\small
\renewcommand{\arraystretch}{1.4}
\setlength{\tabcolsep}{5pt}
\begin{adjustbox}{width=\textwidth}
\begin{tabular}{l|cc|cc|cc|cc|cc|cc}
\midrule
 & \multicolumn{2}{c|}{\textbf{Input Token}}
 & \multicolumn{2}{c|}{\textbf{Output Token}}
 & \multicolumn{2}{c|}{\textbf{Search Count}}
 & \multicolumn{2}{c|}{\textbf{Insert Count}}
 & \multicolumn{2}{c|}{\textbf{Remove Count}}
 & \multicolumn{2}{c}{\textbf{Insert Fail}} \\
\cline{2-13}
 & 1-step & 2-step & 1-step & 2-step & 1-step & 2-step & 1-step & 2-step & 1-step & 2-step & 1-step & 2-step\\
\midrule
\textbf{GPT-4o} & 19368 & 13997 & 664 & 162 & 3.41 & 7.96 & 1.28 & 1.83 & 0.00 & 0.00 & 1.46 & 0.25 \\
\textbf{GPT-4o mini}  & 29705 & 45829 & 786 & 346 & 4.03 & 11.13 & 1.81 & 1.32 & 0.00 & 0.00 & 1.85 & 1.13 \\
\midrule
\textbf{Gemma-3-27b-it} & 73409 & 42486 & 1151 & 244 & 5.65 & 13.54 & 1.82 & 1.29 & 0.00 & 0.00 & 1.87 & 0.72 \\
\textbf{Qwen3-32B} & -- & 23333 & -- & 469 & -- & 4.0 & -- & 3.74 & -- & 0.00 & -- & 1.48 \\
\textbf{Llama-4-Scout-17B-16E-Instruct} & -- & 78488 & -- & 1000 & -- & 12.36 & -- & 0.45 & -- & 0.00 & -- & 5.09 \\
\bottomrule
\end{tabular}
\end{adjustbox}
\end{table*}

\subsection{Tool-Use Efficiency and Self-Correction}
\label{sec:tool_use_analysis}

Table~\ref{tab:count_models} reports average token usage and counts of \texttt{Search}, \texttt{Insert}, and \texttt{Remove} operations.
Across all models and settings, \texttt{Remove} operations are never invoked, indicating limited self-correction during drafting.
This reflects the absence of explicit failure feedback in newswriting tasks, where incorrect drafts are not directly verifiable.
Importantly, this behavior arises from agent decisions rather than framework limitations, as removal is explicitly supported by the interface.
We also observe substantial variation in search--edit efficiency across models.
While the 2-step execution setting increases search attempts, it does not proportionally increase insertions, suggesting broader exploration with reduced integration efficiency.
At the same time, decomposing actions reduces insertion failures, highlighting a trade-off between exploration and reliability.

\subsection{Case Study}
\label{sec:case_study}

This section presents a qualitative comparison illustrating why alignment with human-selected evidence does not necessarily determine end-to-end newswriting quality.
Figure~\ref{fig:qwen3_vs_human_example} compares a BBC human-written article with a Qwen3-32B-generated article for the same event.
The human-written article emphasizes concise factual delivery and balanced reporting.
In contrast, Qwen3-32B incorporates a broader range of historical context and uses it to enhance narrative continuity.
This example illustrates how alternative evidence selection strategies can yield strong end-to-end evaluations despite low alignment with reference articles.

\section{Reproducibility Details}
This section details the implementation.
We first present the complete agent prompt and action interface (Section~\ref{sec:agent_prompt}),
followed by execution settings and API parameters (Section~\ref{sec:exec_settings}),
and the prompt used for end-to-end evaluation (Section~\ref{sec:judge_prompt}).

\subsection{Agent Prompt and Action Specification}
\label{sec:agent_prompt}
We adopt a ReAct-style agentic interface in which agents iteratively alternate between reasoning and tool execution.
The complete agent prompt is presented below in modular blocks for readability; together, these blocks constitute the full system prompt used for all evaluated models.
All models are evaluated under identical prompting and action schemas, ensuring that performance differences reflect agent behavior rather than interface variation.

\begin{tcolorbox}[
  colback=gray!8,
  colframe=gray!50,
  title=Agent Prompt (Part I): Role and High-Level Instructions,
  breakable
]
\label{box:agent_prompt}
\begin{lstlisting}[basicstyle=\ttfamily\small,breaklines=true]
You are a professional news reporter agent that builds a coherent news draft step by step using a strict Thought -> Action loop.

Your task is to construct the draft using only evidence retrieved from prior search observations. You must not introduce information not present in the observations.
\end{lstlisting}
\end{tcolorbox}

\begin{tcolorbox}[
  colback=gray!8,
  colframe=gray!50,
  title=Agent Prompt (Part II): Interaction Protocol,
  breakable
]
\begin{lstlisting}[basicstyle=\ttfamily\small,breaklines=true]
At each step, output only:
- Thought: your reasoning and plan for the next action.
- Action: exactly one action, formatted as JSON.
Do not output any Observation. The system will return the Observation after you submit your Action.
You must wait for the Observation before issuing the next Action.
Only output JSON (start with '{' and end with '}').
\end{lstlisting}
\end{tcolorbox}

\begin{tcolorbox}[
  colback=gray!8,
  colframe=gray!50,
  title=Agent Prompt (Part III): Action Output,
  breakable
]
\begin{lstlisting}[basicstyle=\ttfamily\small,breaklines=true]
{
  "Thought": "<your reasoning>",
  "Action": "<search | insert | remove | terminate>",
  "query": {},
  "year": "",
  "month": "",
  "day": ""
}
\end{lstlisting}
\end{tcolorbox}

\begin{tcolorbox}[
  colback=gray!8,
  colframe=gray!50,
  title=Agent Prompt (Part IV): Constraints,
  breakable
]
\begin{lstlisting}[basicstyle=\ttfamily\small,breaklines=true]
- Only one action is allowed per step.
- Insert actions may only use objects returned by a prior search.
- All inserted content must exactly match retrieved objects.
- Remove actions may only target previously inserted content.
- Continue until issuing a terminate action.
\end{lstlisting}
\end{tcolorbox}

\begin{tcolorbox}[
  colback=gray!8,
  colframe=gray!50,
  title=Agent Prompt (Part V): Available Actions,
  breakable
]
\begin{lstlisting}[basicstyle=\ttfamily\small,breaklines=true]
1. search
- Perform semantic search across the dataset. 
- Required fields: query, year, month, day
Example:
{
  "Thought": "Search for reactions to Sue Gray's report.",
  "Action": "search",
  "query": "Sue Gray report reaction",
  "year": "2022",
  "month": "01",
  "day": "31"
}

2. insert
- Insert one retrieved object into the draft
Example:
{
  "Thought": "Insert a key quote.",
  "Action": "insert",
  "obj": { "text": "Sue Gray's report criticized leadership at No.10." }
}

3. remove
- Remove a previously inserted object
4. terminate
- End the task and finalize the draft
\end{lstlisting}
\end{tcolorbox}

All models are evaluated under identical prompting, action schemas, and operation limits.
Differences in performance therefore reflect model behavior rather than interface variation.

\subsection{Execution Settings and API Parameters}
\label{sec:exec_settings}
All agents are evaluated with default model parameters provided by their respective APIs.
We fix the temperature to 0 to reduce stochastic variation during tool-use and evaluation.
A maximum of 20 actions is allowed as a budget cap, though agents may terminate earlier at their discretion.
For pairwise evaluation, candidate article order is randomized to mitigate positional bias.

\subsection{Evaluation Prompt for End-to-End Comparison}
\label{sec:judge_prompt}
We evaluate end-to-end newswriting quality using a dimension-wise comparison prompt.
The complete evaluation prompt is shown in Box~\ref{box:judge_prompt}.

\begin{tcolorbox}[
  colback=gray!8,
  colframe=gray!50,
  title=Dimension-Wise Evaluation Prompt,
  breakable
]
\label{box:judge_prompt}
\begin{lstlisting}[basicstyle=\ttfamily\small,breaklines=true]
You are an expert evaluator of news articles. 

Evaluate the two candidate articles on **6 dimensions** below. Decide the winner per dimension, then pick an **Overall** winner. Briefly explain each choice. **Return a single JSON object. Do NOT use Markdown code fences or backticks.**

Dimensions:
1.Factual Consistency:factually sound and correct.
2.Logical Consistency:coherent and self-consistent.
3.Importance:conveys more important information.
4.Readability:fluent and easy to read.
5.Objectivity:neutral, minimal opinion.
6.Journalistic Style:adheres to journalistic style.

Overall:best considering all dimensions above. **No tie allowed.**

First Article:
{first}

Second Article:
{second}

Return **only** JSON with this schema (no extra text):

{{
  "Factual Consistency": {{"winner": "first"|"second"|"tie", "reasoning": "brief"}},
  "Logical Consistency": {{"winner": "first"|"second"|"tie", "reasoning": "brief"}},
  "Importance": {{"winner": "first"|"second"|"tie", "reasoning": "brief"}},
  "Readability": {{"winner": "first"|"second"|"tie", "reasoning": "brief"}},
  "Objectivity": {{"winner": "first"|"second"|"tie", "reasoning": "brief"}},
  "Journalistic Style": {{"winner": "first"|"second"|"tie", "reasoning": "brief"}},
  "Overall": {{"winner": "first"|"second", "reasoning": "brief"}}
}}
\end{lstlisting}
\end{tcolorbox}

\section{Human Validation Protocol}
\label{sec:human_validation}

This section describes the human validation procedure used to assess the reliability of the LLM-based judging protocol in our end-to-end evaluation. The goal is not to define an absolute ground truth for news quality, but to measure how closely automatic judgments align with human preferences under a controlled pairwise comparison setting.

\subsection{Sampling and Data}
We randomly sampled 100 human-written news articles from the NEWSAGENT benchmark, covering multiple topical domains. For each sampled event, we formed three candidate articles: the original human-written article, a GPT-4o-generated article, and a GPT-4o-mini-generated article. The model-generated articles were produced under the same agentic setting as the main experiments, including the same tool interface, operation budget, and termination criteria.

This yields three candidate articles per event and 300 pairwise comparisons in total by exhaustively pairing the three candidates. Because all three candidates for a given event are grounded in the same news event and temporal context, the resulting comparisons are controlled for topic and release-date constraints.

\subsection{Annotation Procedure}
Two human annotators were shown two candidate articles at a time and asked to select the better article overall based on journalistic quality. They were instructed to consider overall writing quality, including factors such as factual soundness, clarity, relevance, coherence, and style. The source of each article was hidden, and the left--right order was randomized. Each comparison was independently evaluated by two annotators.

Annotators were allowed to select a tie. If the two annotators disagreed on the winner, or if either annotator selected a tie, the pair was conservatively treated as a human tie and excluded from agreement computation. Following this procedure, 17 of the 300 comparisons were excluded, leaving 283 non-tied comparisons for the final analysis.

\subsection{Agreement Computation}
We compare two automatic judging schemes. The first is a standard single-turn evaluation setup, where the judge directly selects the better article overall in one pass. The second is our dimension-wise evaluation protocol, where the judge first compares the two articles across multiple journalistic dimensions and then synthesizes these judgments into an overall decision.

The exact prompt used for the baseline single-turn evaluation is shown below.

\begin{tcolorbox}[
  colback=gray!8,
  colframe=gray!50,
  title=Baseline Single-Turn Evaluation Prompt,
  breakable,
  fontupper=\footnotesize
]
\begin{lstlisting}[basicstyle=\ttfamily\footnotesize,breaklines=true]
You are an expert evaluator of news articles.
Read the two candidate articles below and decide which one is better overall as a news article.

First Article:
{first}

Second Article:
{second}

Return only a JSON object with the following format:
{
  "winner": "first" | "second",
  "reason": "brief explanation"
}
\end{lstlisting}
\end{tcolorbox}

Agreement is measured by exact match between the automatic judge and human preference on the 283 retained comparisons. We evaluate both protocols with two judge models from different model families. For GPT-4, the standard single-turn evaluation achieves 53\% agreement with human preferences, while the dimension-wise evaluation improves this to 72\%. For Gemini-2.5-Flash, the standard single-turn evaluation achieves 59\% agreement, while the dimension-wise evaluation improves this to 69\%.
Overall, these results show that the dimension-wise protocol aligns better with human judgments than single-turn holistic evaluation across judge families. Based on this validation, we use GPT-4 as the primary judge in the main end-to-end evaluation.

\section{Use of AI Assistants}
\label{sec:ai_assistants}
We used a code editor with generative AI functionalities during code development and manuscript preparation. In writing, this assistance was limited to LaTeX code completion and minor language editing. We also used AI assistants to produce miscellaneous single-use data processing scripts and to assist in figure creation. All AI-assisted outputs were reviewed, verified, and accepted by the authors.

\end{document}